\pdfoutput=1

\documentclass[11pt]{article}

\usepackage[]{acl}

\usepackage{times}
\usepackage{latexsym}

\usepackage{emoji}

\usepackage[T1]{fontenc}

\usepackage[utf8]{inputenc}

\usepackage{microtype}
\usepackage{graphicx}
\usepackage{scalerel} 
\usepackage{nicefrac}
\usepackage{booktabs}

\usepackage{caption}
\usepackage{subcaption}

\usepackage{amsmath,amssymb,mathtools}  
\usepackage{multirow}

\usepackage[utf8]{inputenc}
\usepackage{pgfplots}
\DeclareUnicodeCharacter{2212}{−}
\usepgfplotslibrary{groupplots,dateplot}
\usetikzlibrary{patterns,shapes.arrows}
\pgfplotsset{compat=newest}

\usepackage{enumitem}
\setitemize{noitemsep,topsep=0pt,parsep=0pt,partopsep=0pt}
\setenumerate{noitemsep,topsep=0pt,parsep=0pt,partopsep=0pt}

\newcommand\blfootnote[1]{%
  \begingroup
  \renewcommand\thefootnote{}\footnote{#1}%
  \addtocounter{footnote}{-1}%
  \endgroup
}

\newcommand{\downscale}[1]{{$\scalerel*{#1}{j^2}$}}
\title{AdapLeR: Speeding up Inference by Adaptive Length Reduction}

\author{Ali Modarressi$^{\star\emoji[twitter]{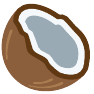}}$ ~ Hosein Mohebbi$^{\star\dagger\emoji[twitter]{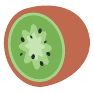}}$ ~ Mohammad Taher Pilehvar$^{\emoji[twitter]{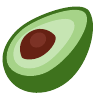}}$ \\
  $^{\emoji[twitter]{coconut}}$ Iran University of Science and Technology, Iran\\
  $^{\emoji[twitter]{kiwi}}$ Cognitive Science and AI, Tilburg University, Netherlands\\
  $^{\emoji[twitter]{avocado}}$ Tehran Institute for Advanced Studies, Khatam University, Iran \\
  \texttt{m\_modarressi@comp.iust.ac.ir}\\
  \texttt{h.mohebbi@uvt.nl}\\ 
  \texttt{mp792@cam.ac.uk}
  }

\begin{document}
\maketitle
\begin{abstract}
Pre-trained language models have shown stellar performance in various downstream tasks.
But, this usually comes at the cost of high latency and computation, hindering their usage in resource-limited settings.
In this work, we propose a novel approach for reducing the computational cost of BERT with minimal loss in downstream performance.
Our method dynamically eliminates less contributing tokens through layers, resulting in shorter lengths and consequently lower computational cost.
To determine the importance of each token representation, we train a Contribution Predictor for each layer using a gradient-based saliency method.
Our experiments on several diverse classification tasks show speedups up to 22x during inference time without much sacrifice in performance.
We also validate the quality of the selected tokens in our method using human annotations in the ERASER benchmark. In comparison to other widely used strategies for selecting important tokens, such as \emph{saliency} and \emph{attention}, our proposed method has a significantly lower false positive rate in generating rationales. Our code is freely available at \url{https://github.com/amodaresi/AdapLeR}.\blfootnote{$^\star$ Equal Contribution.}\blfootnote{$^\dagger$ Work done as a Master's student at IUST.}
\end{abstract}

\section{Introduction}
\label{sec:intro}

While large-scale pre-trained language models exhibit remarkable performances on various NLP benchmarks, their excessive computational costs and high inference latency have limited their usage in resource-limited settings. In this regard,
there have been various attempts at improving the efficiency of BERT-based models \citep{devlin-etal-2019-bert}, including
knowledge distilation \citep{Hinton2015DistillingTK, sanh2019distilbert, sun-etal-2019-patient, sun-etal-2020-mobilebert, jiao-etal-2020-tinybert}, quantization \citep{Gong2014CompressingDC, Shen2020QBERTHB, Tambe2021EdgeBERTSE}, weight pruning \citep{Han2016DeepCC, He2017ChannelPF, Michel2019AreSH, sanh2020movement}, and progressive module replacing \citep{xu-etal-2020-bert}.
Despite providing significant reduction in model size, these techniques are generally static at inference time, i.e., they dedicate the same amount of computation to all inputs, irrespective of their difficulty.

A number of techniques have been also proposed in order to make efficiency enhancement sensitive to inputs.
\emph{Early exit} mechanism \citep{schwartz-etal-2020-right, liao-etal-2021-global, xin-etal-2020-deebert, liu-etal-2020-fastbert, xin-etal-2021-berxit, sun2021early, eyzaguirre2021dact} is a commonly used method in which each layer in the model is coupled with an intermediate classifier to predict the target label. At inference, a halting condition is used to determine whether the model allows an example to exit without passing through all layers.
Various halting conditions have been proposed, including Shannon’s entropy \citep{xin-etal-2020-deebert, liu-etal-2020-fastbert}, softmax outputs with temperature calibration \citep{schwartz-etal-2020-right}, trained confidence predictors \citep{xin-etal-2021-berxit}, or the number of agreements between predictions of intermediate classifiers \citep{zhou2020bert}.

Most of these input-adaptive techniques compress the model from the depth perspective (i.e., reducing the number of involved encoder layers).
However, one can view compression from the width perspective \cite{goyal2020power, ye-etal-2021-tr}, i.e., reducing the length of hidden states. \cite{ethayarajh-2019-contextual, klafka-ettinger-2020-spying}.
This is particularly promising as recent analytical studies showed that there are redundant encoded information in token representations \cite{klafka-ettinger-2020-spying, ethayarajh-2019-contextual}.
Among these redundancies, some tokens carry more task-specific information than others \cite{mohebbi-etal-2021-exploring}, suggesting that only these tokens could be considered through the model.
Moreover, in contrast to layer-wise pruning, token-level pruning does not come at the cost of reducing model's capacity in complex reasoning \cite{sanh2019distilbert, sun-etal-2019-patient}.
PoWER-BERT \citep{goyal2020power} is one of the first such techniques which reduces inference time by eliminating redundant token representations through layers based on self-attention weights. 
Several studies have followed \cite{kim-cho-2021-length, Wang2021SpAttenES};
However, they usually optimize a single token elimination configuration across the entire dataset, resulting in a static model. 
In addition, their token selection strategies are based on attention weights which can result in a suboptimal solution \cite{ye-etal-2021-tr}.

In this work, we introduce \textbf{Adap}tive \textbf{Le}ngth \textbf{R}eduction (\textbf{AdapLeR}).
Instead of relying on attention weights, our method trains a set of Contribution Predictors (CP) to estimate tokens' saliency scores at inference.
We show that this choice results in more reliable scores than attention weights in measuring tokens' contributions.
The most related study to ours is TR-BERT \cite{ye-etal-2021-tr} which leverages reinforcement learning to develop an input-adaptive token selection policy network.
However, as pointed out by the authors, the problem has a large search space, making it difficult for RL to solve. To mitigate this, they resorted to extra heuristics such as imitation learning \cite{hussein2017imitation} for warming up the training of the policy network, action sampling for limiting the search space, and knowledge distillation for transferring knowledge from the intact backbone fine-tuned model. All of these steps significantly increase the training cost. Hence, they only perform token selection at two layers.
In contrast, we propose a simple but effective method to gradually eliminate tokens in each layer throughout the training phase using a soft-removal function which allows the model to be adaptable to various inputs in a batch-wise mode.
It is also worth noting in contrast to our approach above studies are based on top-k operations for identifying the k most important tokens during training or inference, which can be expensive without a specific hardware architecture \cite{Wang2021SpAttenES}.

In summary, our contributions are threefold:
\begin{itemize}
    \item We couple a simple Contribution Predictor (CP) with each layer of the model to estimate tokens' contribution scores to eliminate redundant representations.
    \item Instead of an instant token removal, we gradually mask out less contributing token representations by employing a novel soft-removal function.
    \item We also show the superiority of our token selection strategy over the other widely used strategies by using human rationales.
\end{itemize}

\section{Background}
\subsection{Self-attention Weights}
Self-attention is a core component of the Transformers \cite{NIPS2017_3f5ee243} which looks for the relation between different positions of a single sequence of token representations ($x_1, ..., x_n$) to build contextualized representations. To this end, each input vector $x_i$ is multiplied by the corresponding trainable matrices $Q$, $K$, and $V$ to respectively produce query ($q_i$), key ($k_i$), and value ($v_i$) vectors. 
To construct the output representation $z_i$, a series of weights is computed by the dot product of $q_i$ with every $k_j$ in all time steps.
Before applying a softmax function, these values are divided by a scaling factor and then added to an \emph{attention mask} vector $\mathbf{m}$, which is zero for positions we wish to attend and $-\infty$ (in practice, $-10000$) for padded tokens \cite{NIPS2017_3f5ee243}.
Mathematically, for a single attention head, the weight attention from token $x_i$ to token $x_j$ in the same input sequence can be written as:
\begin{equation}
\label{eq:self_attention}
\alpha_{i,j} = \mathop{\rm{softmax}}_{x_j \in \mathcal{X}}\left(\frac{q_i k_j^\top}{\sqrt{d}} + m_i \right) \in \mathbb{R}
\end{equation}
The time complexity for this is $O(n^2)$ given the dot product $q_i k_j^\top$, where $n$ is the input sequence length.
This impedes the usage of self-attention based models in low-resource settings.

While self-attention is one of the most white-box components in transformer-based models, relying on raw attention weights as an explanation could be misleading given that they are not necessarily responsible for determining the contribution of each token in the final classifier's decision \citep{jain-wallace-2019-attention, serrano-smith-2019-attention, abnar-zuidema-2020-quantifying}. 
This is based on the fact that raw attentions are being faithful to the local mixture of information in each layer and are unable to obtain a global perspective of the information flow through the entire model \citep{pascual-etal-2021-telling}.

\begin{figure*}[t!]
\centering
\includegraphics[width=0.85\textwidth]{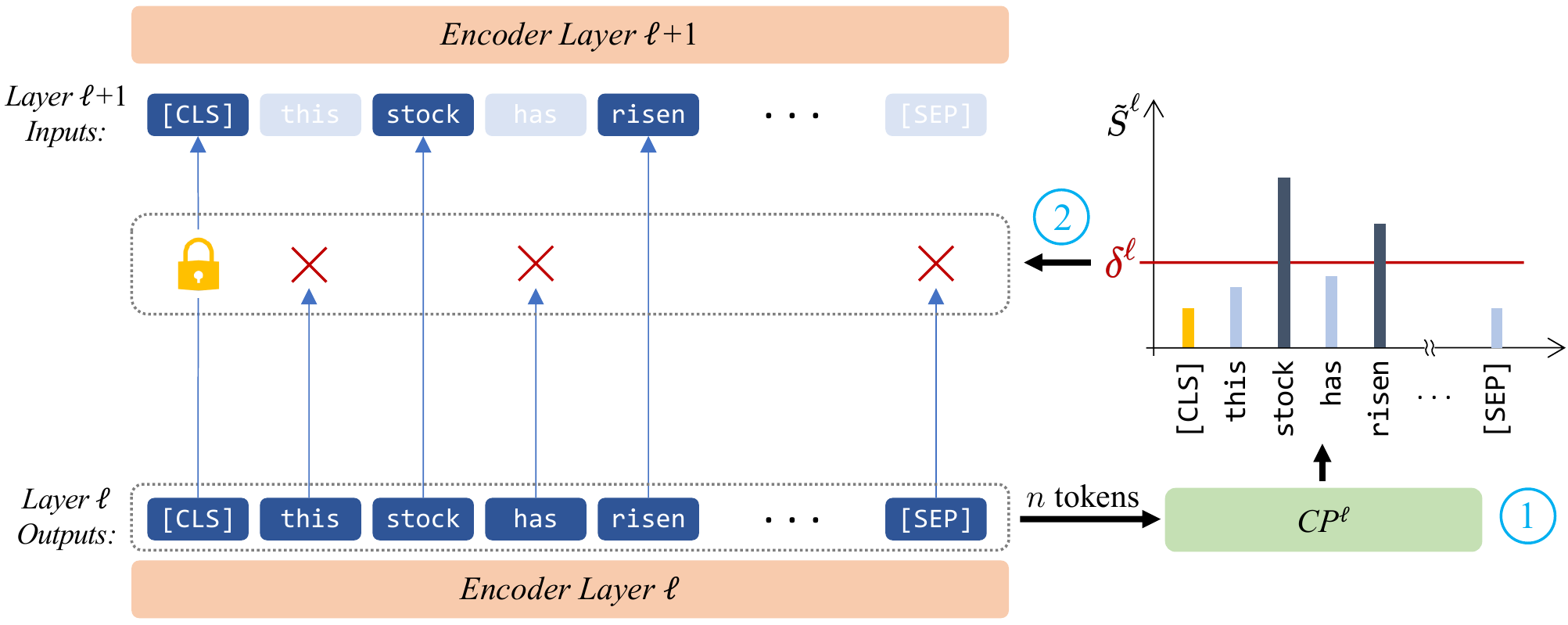} 
\caption{To reduce the inference computation, in each layer (1) the attribution score of the token representation is estimated and (2) based on a reduced uniform-level threshold ($\delta^\ell = \nicefrac{\eta^\ell}{n}$) token representations with low importance score are removed. Since the final layer's classifier is connected to the \textsc{[CLS]} token and it could act as a pooler within each layer it is the only token that would remain regardless of its score.}
\label{fig:lr_method}
\end{figure*}

\subsection{Gradient-based Saliency Scores}
\label{sal_methods}
Gradient-based methods provide alternatives to attention weights to compute the importance of a specific input feature.
Despite having been widely utilized in other fields earlier \citep{ancona2018towards, simonyan2013deep, sundararajan2017axiomatic, smilkov2017smoothgrad}, they have only recently become popular in NLP studies \citep{bastings-filippova-2020-elephant, li-etal-2016-visualizing, yuan2019interpreting}.
These methods are based on computing the first-order derivative of the output logit $y_c$ w.r.t. the input embedding $h^0_i$ (initial hidden states), where $c$ could be true class label to find the most important input features or the predicted class to interpret model's behavior.
After taking the norm of output derivatives, we get \emph{sensitivity} \citep{ancona2018towards}, which indicates the changes in model's output with respect to the changes in specific input dimensions.
Instead, by multiplying gradients with input features, we arrive at \emph{gradient$\times$input} \citep{bastings-filippova-2020-elephant}, 
also known as \emph{saliency}, 
which also considers the direction of input vectors to determine the most important tokens.
Since these scores are computed for each dimension of embedding vectors, an aggregation method such as L2 norm or mean is needed to produce one score per input token \citep{atanasova-etal-2020-diagnostic}:

\begin{equation}
\label{eq:sal_score}
    S_i = \parallel \frac{\partial y_c}{\partial h^0_i} \odot h^0_i \parallel_2
\end{equation}

\section{Methodology}
As shown in Figure \ref{fig:lr_method}, our approach relies on dropping low contributing tokens in each layer and passing only the more important ones to the next. 
Therefore, one important step is to measure the importance of each token. 
To this end, we opted for saliency scores which have been recently shown as a reliable criterion in measuring token's contributions \cite{bastings-filippova-2020-elephant, pascual-etal-2021-telling}.
In Section \ref{sal_vs_attn} we will show results for a series quantitative analyses that supports this choice.
In what follows, we first describe how we estimate saliency scores at inference time using a set of Contribution Predictors (CPs) and then elaborate on how we leverage these predictors during inference (Section \ref{sec:model_inference}) and training (Section \ref{sec:model_training}).

\subsection{Contribution Predictor}
Computing gradients during inference is problematic as backpropagation computation prolongs inference time, which is contrary to our main goal. 
To circumvent this, we simply add a CP after each layer $\ell$ in the model to estimate contribution score for each token representation, i.e., \downscale{\tilde{S}_i^\ell}.
The model then decides on the tokens that should be passed to the next layer based on the values of \downscale{\tilde{S}_i^\ell}.
CP computes \downscale{\tilde{S}_i^\ell} for each token using an MLP followed by a softmax activation function.
We argue that, despite being limited in learning capacity, the MLP is sufficient for estimating scores that are more generalized and relevant than vanilla saliency values. 
We will present a quantitative analysis on this topic in Section \ref{sec:analysis}.

\subsection{Model Inference}
\label{sec:model_inference}
Most BERT-based models consist of $L$ encoder layers.
The input sequence of $n$ tokens is usually passed through an embedding layer to build the initial hidden states of the model $h^0$. 
Each encoder layer then produces the next hidden states using the ones from the previous layer:
\begin{equation}
\label{eq:encoder}
    h^\ell = \text{Encoder}_\ell(h^{\ell-1})
\end{equation}

In our approach, we eliminate less contributing token representations before delivering hidden states to the next encoder.
Tokens are selected based on the contribution scores \downscale{\tilde{S}^\ell} obtained from the CP of the corresponding layer $\ell$.
As the sum of these scores is equal to one, a uniform level indicates that all tokens contribute equally to the prediction and should be retained. On the other hand, the lower-scoring tokens could be viewed as unnecessary tokens if the contribution scores are concentrated only on a subset of tokens. 
Given that the final classification head uses the last hidden state of the \textsc{[CLS]} token, we preserve this token's representation in all layers. Despite preserving this, other tokens might be removed from a layer when \textsc{[CLS]} has a significantly high estimated contribution score than others. Based on this intuition, we define a cutoff threshold based on the uniform level as: $\delta^\ell = \eta^\ell\cdot\nicefrac{1}{n}$ with $0 < \eta^\ell \leq 1$ to distinguish important tokens.
Tokens are considered important if their contribution score exceeds $\delta$ (which is a value equal or smaller than the uniform score).
Intuitively, a larger $\eta$ provides a higher $\delta$ cutoff level, thereby dropping a larger number of tokens, hence, yielding more speedup. 
The value of $\eta$ determines the extent to which we can rely on CP's estimations. 
In case the estimations of CP are deemed to be inaccurate, its impact can be reduced by lowering $\eta$.
We train each layer's $\eta^\ell$ using an auxiliary training objective, which allows the model to adjust the cutoff value to control the speedup-performance tradeoff. 
Also, since each input instance has a different computational path during token removal process, it is obvious that at inference time, the batch size should be equal to one (single instance usage), similarly to other dynamic approaches \citep{zhou2020bert, liu-etal-2020-fastbert, ye-etal-2021-tr, eyzaguirre2021dact, xin-etal-2020-deebert}. 

\subsection{Model Training}
\label{sec:model_training}
Training consists of three phases: initial fine-tuning, saliency extraction, and adaptive length retraining. 
In the first phase, we simply fine-tune the backbone model (BERT) on a given target task. 
We then extract the saliencies of three top-perfroming checkpoints from the fine-tuning process and compute the average of them to mitigate potential inconsistencies in saliency scores (cf. Section \ref{sal_methods}).
The final step is to train a pre-trained model using an adaptive length reduction procedure. 
In this phase, a non-linear function gradually fades out the representations throughout the training process. Each CP is jointly trained with the rest of the model using the saliencies extracted in the previous phase alongside with the target task labels. 
We also define a speedup tuning objective to determine the thresholds (via tuning $\eta$) to control the performance-speedup trade-off.
In the following, we elaborate on the procedure.

\begin{figure}[t!]
\centering
\includegraphics[width=\linewidth]{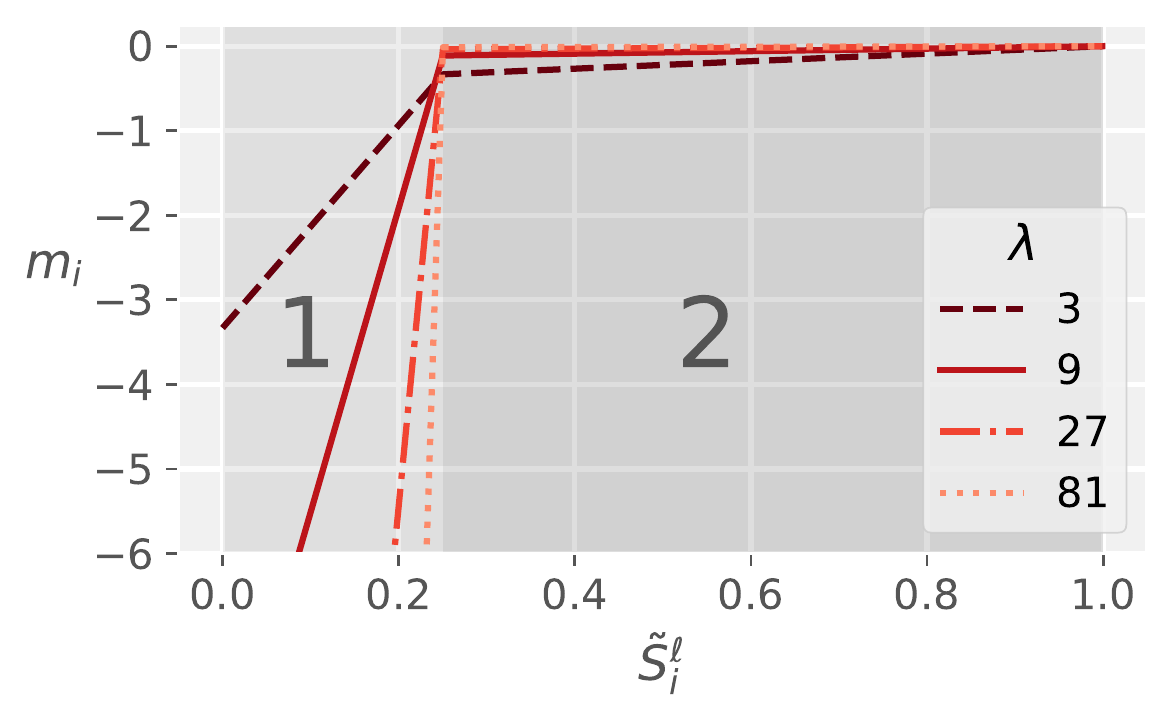}
\caption{The soft-removal function plotted with \mbox{$\lambda \in \{3,9,27,81\}$} and $\delta^\ell = 0.25$. As $\lambda$ increases, the removal region (1) gets steeper while the other zone (2), which is almost horizontal, approaches the zero level.}
\label{fig:soft-removal}
\end{figure}

\paragraph{Soft-removal function.}
During training, if tokens are immediately dropped similarly to the inference mode, the effect of dropping tokens cannot be captured using a gradient backpropagation procedure. Using batch-wise training in this scenario will also be problematic as the structure will vary with each example.
Hence, inspired by the padding mechanism of self-attention models \mbox{\citep{NIPS2017_3f5ee243}} we introduce a new procedure that gradually masks out less contributing token representations. 
In each layer, after predicting contribution scores, instead of instantly removing the token representations, we accumulate a negative mask to the attention mask vector $M$ using a soft-removal function:
\begin{equation}
\label{eq:soft-removal}
    m_i^{-}(\tilde{S}_i^\ell)=
    \begin{dcases}
        \lambda_{adj}(\tilde{S}_i^\ell - \delta^\ell) - \frac{\beta}{\lambda} & \tilde{S}_i^\ell < \delta^\ell\\
        \frac{(\tilde{S}_i^\ell - 1)\beta}{(1 - \delta^\ell)\lambda}              & \tilde{S}_i^\ell \geq \delta^\ell
    \end{dcases}
\end{equation}
This function consists of two main zones (Figure \ref{fig:soft-removal}).
In the first term, the less important tokens with scores lower than the threshold ($\delta^\ell$) are assigned higher negative masking as they get more distant from $\delta$. 
The slope is determined by $\lambda_{adj}=\nicefrac{\lambda}{\delta}$, where $\lambda$ is a hyperparameter that is increased exponentially after each epoch (e.g., \mbox{$\lambda\leftarrow10\times\lambda$} after finishing each epoch).
Increasing $\lambda$ makes the soft-removal function stronger and more decisive in masking the representations. 
To avoid undergoing zero gradients during training, we define $0 < \beta < 0.1$ to construct a small negative slope (similar to the well known Leaky-ReLU of \citealt{Maas13rectifiernonlinearities}) for those tokens with higher contributing scores than $\delta^\ell$ threshold.
Consider a scenario in which $\eta^\ell$ sharply drops, causing most of \downscale{\tilde{S}^\ell_i} get over the $\delta^\ell$ threshold. In this case, the non-zero value in the second term of Equation \ref{eq:soft-removal}, which facilitates optimizing $\eta^\ell$.

\paragraph{Training the Contribution Predictors.}
The CPs are trained by an additional term which is based on the KL-divergence\footnote{Inclusive KL loss. Check \mbox{Appendix \ref{sec:app_kl_loss}.}} of each layer's CP output with the extracted saliencies. The main training objective is a minimization of the following loss:
\begin{equation}
\label{eq:loss_1}
\mathcal{L} = \mathcal{L}_\textsc{CE} + \gamma\mathcal{L}_\textsc{CP}
\end{equation}
Where $\gamma$ is a hyperparameter which that specifies the amount of emphasis on the CP training loss:
\begin{equation}
\label{eq:loss_CP}
\begin{aligned}
\mathcal{L}_\textsc{CP} &= \sum_{\ell=0}^{L-1}(L-\ell)D_\textsc{KL}(\hat{S}^\ell || \tilde{S}^\ell) \\
&= \sum_{\ell=0}^{L-1}(L-\ell)\sum_{i=1}^{N}\hat{S}^\ell_i\log(\frac{\hat{S}^\ell_i}{\tilde{S}^\ell_i})
\end{aligned}
\end{equation}
Since $S$ is based on the input embeddings, the \textsc{[CLS]} token usually shows a low amount of contribution due to not having any contextualism in the input.
As we leverage the representation of the  \textsc{[CLS]} token in the last layer for classification, this token acts as a pooler and gathers information about the context of the input. In other words, the token can potentially have more contribution as it passes through the model. 
To this end, we amplify the contribution score of \textsc{[CLS]} and renormalize the distribution ($\hat{S}^\ell$) with a trainable parameter $\theta^\ell$:
\begin{equation}
\label{eq:S_hat} 
\hat{S}_i^\ell=\frac{\theta^\ell S_1^\ell\mathbf{1}[i=1] + S_i^\ell\mathbf{1}[i>1]}{\theta^\ell S_1^\ell+\sum_{i=2}^{n}S_i^\ell}
\end{equation}

By this procedure, the next objective (discussed in the next paragraph) will have the capability of tuning the amount of pooling, consequently controlling the amount of speedup. 
Larger $\theta$ push the CPs to shift the contribution towards the \textsc{[CLS]} token to gather most of the task-specific information and avoids carrying redundant tokens through the model.

\paragraph{Speedup Tuning.}
In the speedup tuning process, we combine the cross-entropy loss of the target classification task with a length loss which is the expected number of unmasked token representations in all layers. Considering that we have a non-positive and continuous attention mask $M$, the length loss of a single layer would be the summation over the exponential of the mask values $\exp(m_i)$ to map the masking range $[-\infty, 0]$ to a $[0\text{ (fully masked/removed)}, 1\text{ (fully retained)}]$ bound.
\begin{equation}
\label{eq:loss_2}
\begin{aligned}
\mathcal{L}_\textsc{Spd./Perf.} &= \mathcal{L}_\textsc{CE} + \phi \mathcal{L}_\textsc{Length} \\
\mathcal{L}_\textsc{Length} &= \sum_{l=1}^{L}\sum_{i=1}^{n}\exp({m_i^\ell})
\end{aligned}
\end{equation}
Equation \ref{eq:loss_2} demonstrates how the length loss is computed inside the model and how it is added to the main classification loss. During training, we assign a separate optimization process which tunes $\eta$ and $\theta$ to adjust the thresholds and the amount of \textsc{[CLS]} pooling\footnote{Since $\theta$ is not in the computational DAG, we employed a dummy variable inside the model.
See Appendix \ref{sec:app_theta_optimization}.} alongside with the CP training. 

The reason that this objective is treated as a separate problem instead of merging it with the previous one, is because in the latter case the CPs could be influenced by the length loss and try to manipulate the contribution scores for some tokens regardless of their real influence. So in other words, the first objective is to solve the task and make it explainable with the CPs, and the secondary objective builds the speedup using tuning the threshold levels and the amount of pooling in each layer.

\begin{table*}[!t]  
\resizebox{\textwidth}{!}{
\centering
\setlength{\tabcolsep}{4pt}
\begin{tabular}{l r r r r r r r r r r r r r r r r}
\toprule

\multirow{2}{*}{Model}   & \multicolumn{2}{c}{SST-2} & \multicolumn{2}{c}{IMDB} & \multicolumn{2}{c}{HateXplain} & \multicolumn{2}{c}{MRPC} &  \multicolumn{2}{c}{MNLI} & \multicolumn{2}{c}{QNLI} & \multicolumn{2}{c}{AG’s news} & \multicolumn{2}{c}{DBpedia}\\
\cmidrule(lr){2-3}
\cmidrule(lr){4-5}
\cmidrule(lr){6-7}
\cmidrule(lr){8-9}
\cmidrule(lr){10-11}
\cmidrule(lr){12-13}
\cmidrule(lr){14-15}
\cmidrule(lr){16-17}

& Acc. & Speedup & Acc. & Speedup & Acc & Speedup & F1. & Speedup & Acc. & Speedup & Acc. & Speedup & Acc. & Speedup & Acc. & Speedup \\
\midrule
BERT & 92.7 & 1.00x~~~ & 93.8 & 1.00x~~~ & 68.3 & 1.00x~~~ & 87.5 & 1.00x~~~ & 84.2 & 1.00x~~~ & 90.3 & 1.00x~~~ & 94.4 & 1.00x~~~ & 99.3 & 1.00x~~~ \\
\midrule
DistilBERT & 92.2 & 2.00x~~~ & 92.9 & 2.00x~~~ & 68.2 & 2.00x~~~ & 88.0 & 2.00x~~~ & 81.8 & 2.00x~~~ & 88.1 & 2.00x~~~ & 94.2 & 2.00x~~~ & 99.3 & 2.00x~~~ \\
\midrule
PoWER-BERT & 92.1 & 1.18x~~~ & 92.2 & 1.70x~~~ & 66.9 & 2.69x~~~ & 88.0 & 1.07x~~~ & 82.9 & 1.10x~~~ & 89.7 & 1.23x~~~ & 92.1 & 12.50x~~~ & 98.1 & 14.80x~~~ \\
TR-BERT & 92.1 & 1.46x~~~ & 93.2 & 2.90x~~~ & 67.9 & 2.23x~~~ & 81.9 & 1.16x~~~ & 84.8 & 1.00x~~~ & 89.0 & 1.09x~~~ & 93.2 & 10.20x~~~ & 98.9 & 10.01x~~~ \\
\midrule
\textbf{AdapLeR} & 92.3 & 1.49x~~~ & 91.7 & 3.21x~~~ & 68.6 & 4.73x~~~ & 87.6 & 1.27x~~~ & 82.9 & 1.42x~~~ & 89.3 & 1.47x~~~ & 92.5 & 17.10x~~~ & 98.9 & 22.23x~~~ \\
\bottomrule
\end{tabular}
}
\caption{Comparison of our proposed method (AdapLeR) with other baselines in eight classification tasks in terms of performance and speedup. For each dataset the corresponding metric has been reported (Accuracy: Acc., F1: F-1 Score). In the MNLI task, the speedup and performance values are the average of the evaluations on the matched and mismatched test sets.
}
\label{tab:results}
\end{table*}

\section{Experiments}
\subsection{Datasets}
To verify the effectiveness of AdapLeR on inference speedup, we selected eight various text classification datasets. 
In order to incorporate a variety of tasks, we utilized SST-2 \citep{socher-etal-2013-recursive-sst2} and IMDB \citep{maas-etal-2011-learning_imdb} for sentiment, MRPC \citep{dolan-brockett-2005-automatically-mrpc} for paraphrase, AG's News \citep{Zhang2015CharacterlevelCN_agnews} for topic classification, DBpedia \citep{lehmann2015dbpedia} for knowledge extraction, MNLI \citep{williams-etal-2018-broad-mnli} for NLI, QNLI \citep{rajpurkar-etal-2016-squad-qnli} for question answering, and \mbox{HateXplain \citep{mathew2021hatexplain}} for hate speech.\footnote{See the statistics of datasets in Table \ref{tab:data_statitics} in Appendix.}
Evaluations are based on the test split of each dataset. For those datasets that are in the GLUE Benchmark \citep{wang-etal-2018-glue}, test results were acquired by submitting the test predictions to the evaluation server.

\subsection{Experimental Setup}
As our baseline, we report results for the pre-trained BERT model (base-uncased) \citep{devlin-etal-2019-bert} which is also the backbone of AdapLeR.
We also compare against three other approaches: \mbox{DistilBERT (uncased)} \citep{sanh2019distilbert} as a static compression method, PoWER-BERT and TR-BERT as two strong length reduction methods (cf. Sec. \ref{sec:intro}). 
We used the provided implementations and suggested hyperparameters\footnote{Since some of the datasets were not used originally, we had to search the hyperparameters based on the given ranges.} to train these baselines.
To fine-tune the backbone model, we used same hyperparameters over all tasks (see Section \ref{sec:Hyperparams_app} for details).
The backbone model and our model implementation is based on the HuggingFace's Transformers library \citep{wolf-etal-2020-transformers}. 
Trainings and evaluations were conducted on a dual 2080Ti 11GB GPU machine with multiple runs.

\paragraph{Hyperparameter Selection.}
Overall, we introduced four hyperparameters ($\gamma, \phi, \lambda, \beta$)\footnote{Note that $\theta$ and $\eta$ are trainable terms that are tuned by the model during training.} which are involved in the training process. 
Among these, $\phi$ and $\gamma$ are the primary terms that have considerable effects on AdapLeR's downstream performance and speedup.
This makes our approach comparable to existing techniques \citep{goyal2020power, ye-etal-2021-tr} which usually have two or three hyperparameters adjusted per task.
We used grid search to find the optimal values for these two terms, while keeping the other hyperparameters constant over all datasets. 
Hyperparamter selection is further discussed in Section \ref{sec:Hyperparams_app}.

\paragraph{FLOPs Computation.}
We followed \citet{ye-etal-2021-tr} and \citet{liu-etal-2020-fastbert} and measured computational complexity in terms of FLOPs, i.e., the number of floating-point operations (FLOPs) in a single inference procedure. 
This allows us to assess models' speedups independently of their operating environment (e.g., CPU/GPU).
The total FLOPs of a given model is a summation of the measured FLOPs over all test examples. 
Then, a model's speedup can be defined as the total FLOPs measured on BERT (our baseline) divided by the corresponding model's total FLOPs.
To have a fair comparison, we also computed FLOPs for PoWER-BERT in a single instance mode, described in Section \ref{sec:power-bert-single-instance}.

\begin{figure}[!t]
\centering
     \begin{subfigure}[b]{\linewidth}
         \centering
         \includegraphics[width=0.81\linewidth]{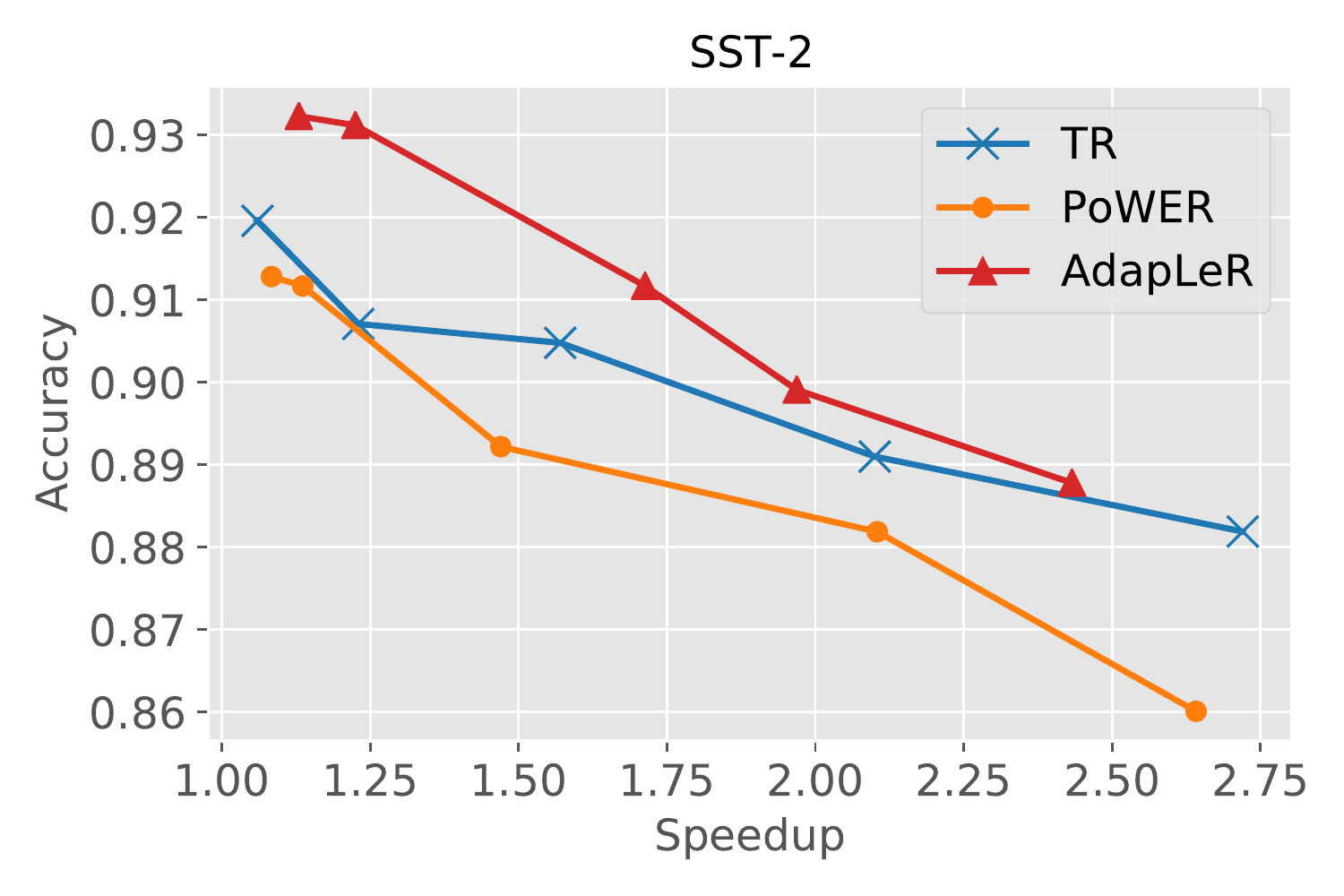}
     \end{subfigure}
     \begin{subfigure}[b]{\linewidth}
         \centering
         \includegraphics[width=0.81\linewidth]{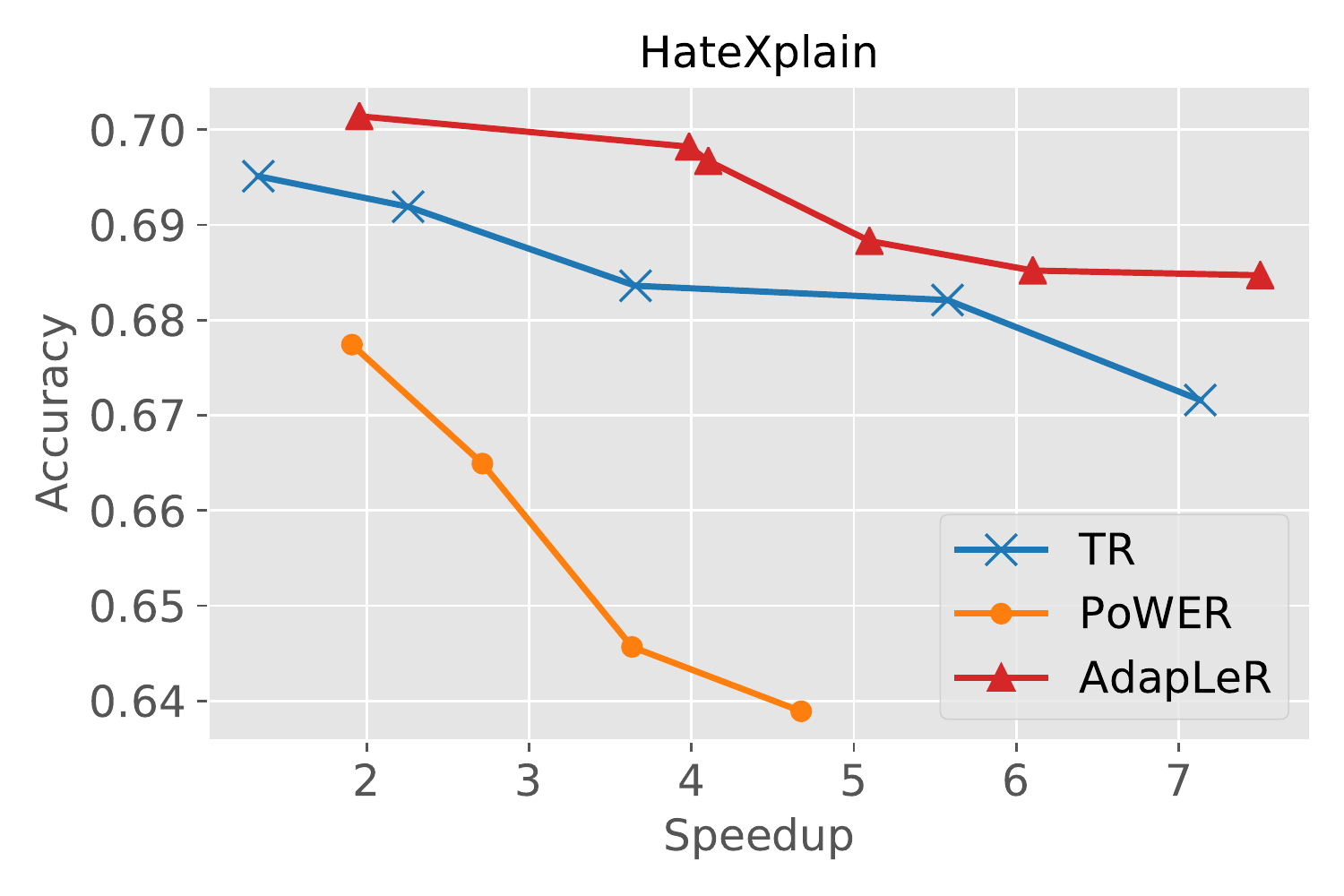}
     \end{subfigure}
        \caption{Accuracy-Speedup trade-off curve for AdapLeR and two other state-of-the-art reduction methods; TR: TR-BERT, PoWER: PoWER-BERT on two different tasks.}
        \label{fig:pareto}
\end{figure}

\subsection{Results}
\label{sec:main_results}
Table \ref{tab:results} shows performance and speedup for AdapLeR and other comparison models across eight different datasets. 
While preserving the same level of performance, AdapLeR outperforms other techniques in terms of speedup across all tasks (ranging from +0.2x to +7.4x compared to the best model in each dataset).

It is noteworthy that the results also reveal some form of dependency on the type of the tasks.
Some tasks may need less amount of contextualism during inference and could be classified by using only a fraction of input tokens. 
For instance, in AG's News, the topic of a sentence might be identifiable with a single token (e.g., \mbox{\emph{soccer} $\,\to\,$ Topic: Sports,} see Figure \ref{fig:additional_qualitative_analysis} in the Appendix for an example).
PoWER-BERT adopts attention weights in its token selection which requires at least one layer of computation to be determined, and TR-BERT applies token elimination only in two layers to reduce the training search space. 
In contrast, our procedure performs token elimination for all layers of the model, enabling a more effective removal of redundant tokens. 
On the other hand, we observe that TR-BERT and PoWER-BERT lack any speedup gains for tasks such as QNLI, MNLI, and MRPC which need a higher degree of contextualism during inference.
However, AdapLeR can offer some speedups even for these tasks.

\paragraph{Speedup-Performance Tradeoff.} 
To provide a closer look at the efficiency of AdapLeR in comparison with the other state-of-the-art length reduction methods, we illustrate speedup-accuracy curves in Figure \ref{fig:pareto}. 
We provide these curves for two tasks in which other length reduction methods show comparable speedups to AdapLeR.
For each curve, the points were obtained by tuning the most influential hyperparameters of the corresponding model.
As we can see, AdapLeR significantly outperforms the other two approaches in all two tasks.
An interesting observation here is that the curves for TR-BERT and AdapLeR are much higher than that of PoWER-BERT. 
This can be attributed to the input-adaptive procedure employed by the former two methods for determining the number of reduced tokens (whereas PoWER-BERT adopts a fixed retention configuration in token elimination).

\section{Analysis}
\label{sec:analysis}
In this section, we first conduct an experiment to support our choice of saliency scores as a supervision in measuring the importance of token representations. 
Next, we evaluate the behavior of Contribution Predictors in identifying the most important tokens in the AdapLeR.

\begin{figure*}[t!]
\centering
    \includegraphics[width=0.95\linewidth]{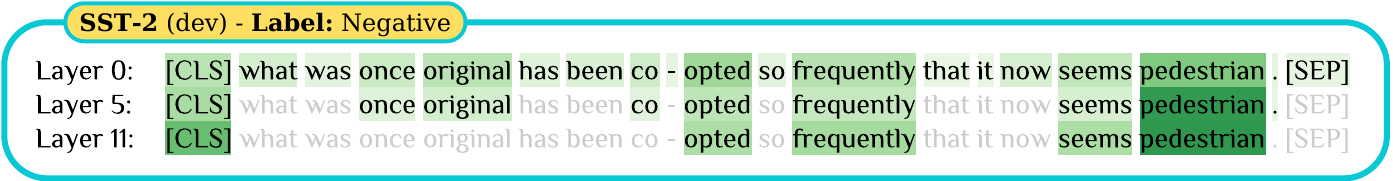}
    \includegraphics[width=0.95\linewidth]{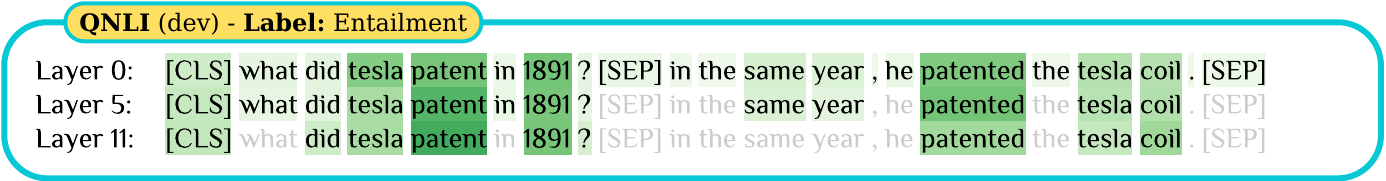}
    \caption{The illustration of contribution scores obtained by CPs in three different layers of the model for two input examples from SST-2 (sentiment) and QNLI (Question-answering NLI) tasks. The contribution scores are shown by color intensity. Only the highlighted token representations are processed in each layer. See more full-layer plots in the appendix \ref{fig:additional_qualitative_analysis}.}
    \label{fig:qualitative_analysis}
\end{figure*}

\subsection{Rationale as an Upper Bound}
\label{sal_vs_attn}

A natural question that arises when dealing with token pruning is that of \emph{importance measure}: what is the most appropriate criterion for assessing the relative importance of tokens within a sentence?
We resort to human rationale as a reliable upper bound for measuring token importance.
To this end, we used the ERASER benchmark \citep{deyoung-etal-2020-eraser}, which contains multiple tasks for which important spans of the input text have been highlighted as supporting evidence (aka ``rationale'') by human.
Among the tasks in the benchmark, we opted for two diverse classification tasks: Movie reviews \citep{zaidan-eisner-2008-modeling} and MultiRC \citep{khashabi-etal-2018-looking}.
In the former task, the model predicts the sentiment of the passage. Whereas the latter contains a passage, a question, and multiple candidate answers, which is cast as a binary classification task of passage/question/answer triplets in the ERASER benchmark.

\begin{table}
    \centering
    {
    \resizebox{\columnwidth}{!}
{
    	\begin{tabular}{lcccc}
    		\toprule
    	
    		& \multicolumn{2}{c}{Movie Reviews} & \multicolumn{2}{c}{MultiRC} \\
    		\cmidrule(lr){2-3}
    		\cmidrule(lr){4-5}
    		\textbf{Strategy} & Acc. & Speedup & Acc. & Speedup\\
    		\midrule
    		
    		Full input & 93.3 & 1.0x & 67.7 & 1.0x \\
    		Human rationale & 96.7 & 3.7x & 76.6 & 4.6x \\ 
    		\midrule
    		Saliency & \textbf{92.3} & 3.7x & \textbf{66.4} & 4.4x \\    
    		Attention \textsc{all} & 78.5 & 3.7x & 62.9 & 4.4x \\    
    		Attention \textsc{[CLS]} & 70.3 & 3.7x & 63.7 & 4.4x \\    
    		\bottomrule
    	\end{tabular}
    }
}
    \caption{\label{tab:pilot_analysis} Accuracy and speedup when the most important input tokens during fine-tuning are computed based on attention and saliency strategies and human rationale (the upper bound).
    The bold values indicate the best corresponding strategy for each task (the closest performance to the upper bound).}
\end{table}

In order to verify the reliability of human rationales, we fine-tuned BERT based on the rationales only, i.e., by excluding those tokens that are not highlighted as being important in the input.
In Table \ref{tab:pilot_analysis}, the first two rows show the performance of BERT on the two tasks with full input and with human rationales only.
We see that fine-tuning merely on rationales not only yields less computation cost, but also results in a better performance when compared with the full input setting. 
Obviously, human annotations are not available for a whole range of downstream NLP tasks; therefore, this criterion is infeasible in practice and can only be viewed as an upper bound for evaluating different strategies in measuring token importance.

\subsection{Saliency vs. Attention}
\label{sec:sal_vs_attn}
We investigated the effectiveness of saliency and self-attention weights as two commonly used strategies for measuring the importance of tokens in pre-trained language models.
To compute these, we first fine-tuned BERT with all tokens in the input for a given target task. 
We then obtained saliency scores with respect to the tokens in the input embedding layer. 
This brings about two advantages. 
Firstly, representations in the embedding layer are non-contextualized, allowing us to measure the importance of each token independently from the others.
Secondly, the backpropagation of gradients through layers to the beginning of the model provides us with aggregated values for the relative importance of each token based on the entire model.
Similarly, we aggregated the self-attention weights across all layers of the model using a post-processed variant of attentions called \emph{attention rollout} \citep{abnar-zuidema-2020-quantifying}, a popular technique in which the attention weight matrix in each layer is multiplied with the preceding ones to form aggregated attention values.

To assign an importance score to each token, we examined two different interpretation of attention weights.
The first strategy is the one adopted by PoWER-BERT \citep{goyal2020power} in which for each token we accumulate attention values from other tokens.
Additionally, we measured how much the \textsc{[CLS]} token attends to each token in the sentence, a strategy which has been widely used in interpretability studies around BERT \citep[\textit{inter alia}]{abnar-zuidema-2020-quantifying, Chrysostomou2021EnjoyTS, Jain2020LearningTF}.
For a fair comparison, for each sentence in the test set, we selected the top-$k$ salient and attended words, with $k$ being the number of words that are annotated as rationales.

Results in Table \ref{tab:pilot_analysis} show that fine-tuning on the most salient tokens outperforms that based on the most attended tokens.
This denotes that saliency is a better indicator for the importance of tokens.
Nonetheless, recent length reduction techniques \citep{goyal2020power, kim-cho-2021-length, Wang2021SpAttenES} have mostly adopted attention weights as their criterion for selecting important tokens. Computing these weights is convenient as they are already computed during the forward pass, whereas computing saliency requires an additional backpropagation step. Note that in our approach, saliency scores are easily estimated within inference time by the pre-trained CPs.

\subsection{Contribution Predictor Evaluation}

In this section we validate our Contribution Predictors in selecting the most contributed tokens. 
Figure \ref{fig:qualitative_analysis} illustrates two examples from the SST-2 and QNLI datasets in which CPs identify and gradually drop the irrelevant tokens through layers, finally focusing mostly on the most important token representations; \emph{pedestrian} (adjective) in SST-2 and \emph{tesla coil} in the passage part of QNLI (both of which are highly aligned with human rationale).

\begin{figure}[!t]
    \includegraphics[width=\linewidth]{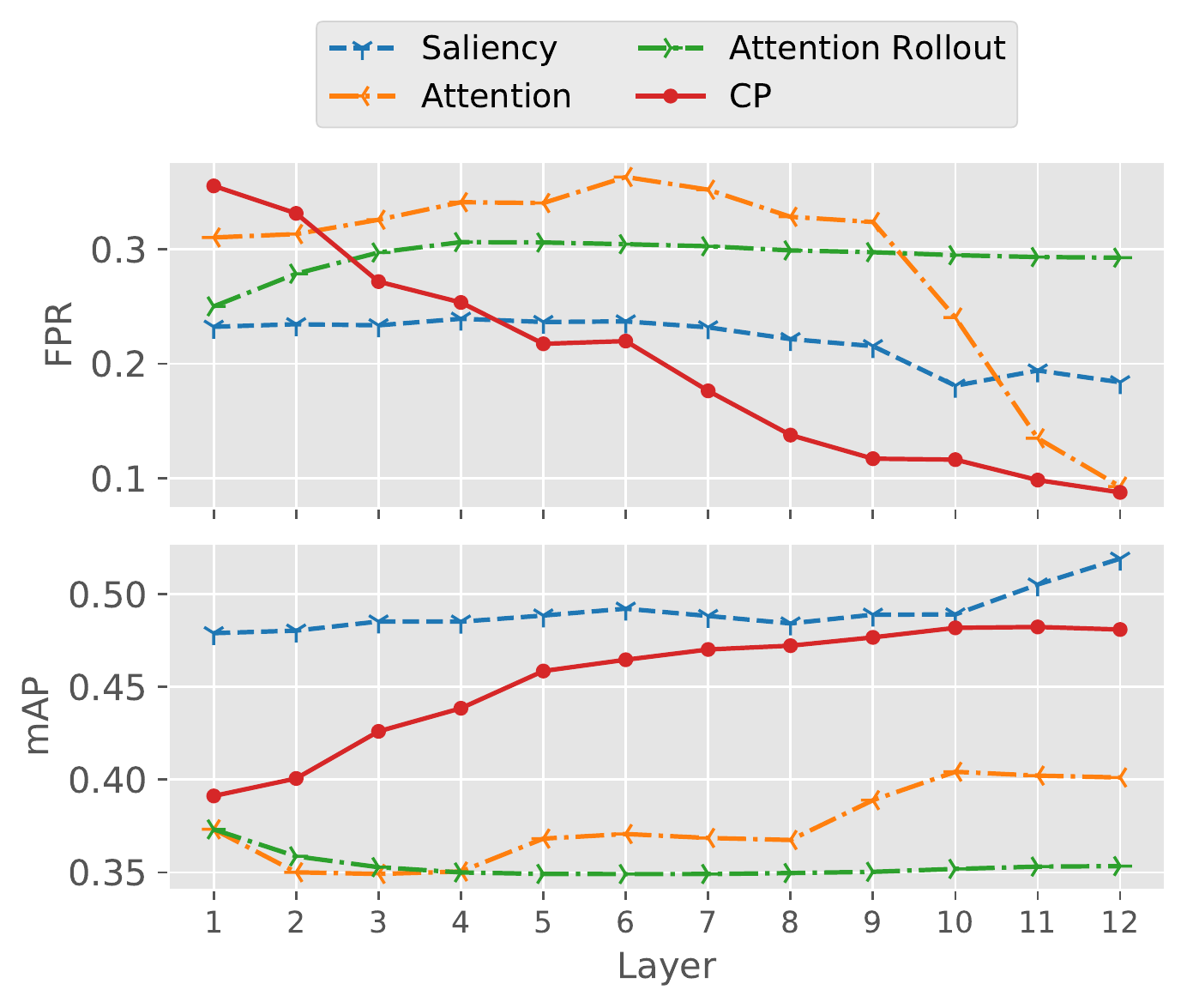}
    \caption{Agreement with human rationales in terms of mean Average Precision and False Positive Rate for Contribution Predictor (CP) and three alternative techniques.}
    \label{fig:q_analysis}
\end{figure}

In order to quantify the extent to which AdapLeR's CPs can preserve rationales without requiring direct human annotations in an unsupervised manner we carried out the following experiment. To investigate the effectiveness of trained CPs in predicting human rationales we computed the output scores of CPs in AdapLeR for each token representation in each layer.
We also fine-tuned a BERT model on the \mbox{Movie Review} dataset and computed layer-wise raw attention, attention rollout, and saliency scores for each token representation.
Since human rationales are annotated at the word level, we sum the scores across tokens corresponding to each word to arrive at word-level importance scores.
In addition to these soft scores, we used the uniform-level threshold (i.e., $\nicefrac{1}{n}$) to reach a binary score indicating tokens selected in each layer.

As for evaluation, we used the Average Precision (AP) and False Positive Rate (FPR) metrics by comparing the remaining tokens to the human rationale annotations. 
The first metric measures whether the model assigns higher continuous scores to those tokens that are annotated by humans as rationales.
Whereas, the intuition behind the second metric is how many irrelevant tokens are selected by the model to be passed to subsequent layers.
We used soft scores for computing AP and binary scores for computing FPR.

Figure \ref{fig:q_analysis} shows the agreement between human rationales and the selected tokens based on the two metrics. 
In comparison with the other widely used strategies for selecting important tokens, such as salinecy and attention, our CPs have significantly less false positive rate in preserving rationales through layers. 
Despite having similar FPRs at the final layer, CP is preferable to attention in that it can better identify rationales at the earlier layers, allowing the model to combine the most relevant token representations when building the final one.
This in turn results in better performance, as was also shown in the previous experiment in Section \ref{sec:sal_vs_attn}.
Also, we see that the curve of mAP for saliency is consistently higher than other strategies in terms of alignment with human rationales which supports our choice of saliency as a measure for token importance.

Finally, we note that there is a line of research that attempts at guiding models to perform human-like reasoning by training rationale generation simultaneously with the target task that requires human annotation \cite{atanasova-etal-2020-generating-fact, Zhao2020Transformer-XH:, li-etal-2018-end}. 
As a by-product of the contribution estimation process, our trained CPs are able to generate these rationales at inference without the need for human-generated annotations.

\section{Conclusion}

In this paper, we introduced AdapLeR, a novel method that accelerates inference by dynamically identifying and dropping less contributing token representations through layers of BERT-based models.
Specifically, AdapLeR accomplishes this by training a set of Contribution Predictors based on saliencies extracted from a fine-tuned model and applying a gradual masking technique to simulate input-adaptive token removal during training. 
Empirical results on eight diverse text classification tasks show considerable improvements over existing methods. 
Furthermore, we demonstrated that contribution predictors generate rationales that are highly in line with those manually specified by humans. 
As future work, we aim to apply our technique to more tasks and see whether it can be adapted to those tasks that require all token representations to be present in the final layer of the model (e.g., question answering). 
Additionally, combining our width-based strategy with a depth-based one (e.g., early exiting) might potentially yield greater efficiency, something we plan to pursue as future work.

\section*{Broader Impact}
Using our proposed method, pre-trained language models can use fewer FLOPs, reducing energy use and carbon emissions \citep{Schwartz2020GreenA}.

\bibliography{anthology,custom}
\bibliographystyle{acl_natbib}

\appendix
\section{Inclusive KL Loss Consideration}
\label{sec:app_kl_loss}
We opted for an inclusive KL loss since CPs should be trained to cover all tokens considered important by saliency and not to be mode seeking (i.e., covering a subset of high contributing tokens considered by the saliency scores.). Suppose an exclusive KL is selected. Due to the limited learning capacity of the CP and miscalculation possibility from the saliency, the CP may be trained to maximize its contribution on noninformative tokens. While in an inclusive setting, it trains to extend its coverage over all high-saliency tokens.

Additionally, our initial research indicated that using a symmetric loss (e.g. Jensen-Shannon divergence) would produce similar results but with a significantly longer convergence time.

\section{Optimization of $\theta$}
\label{sec:app_theta_optimization}
In Section \ref{sec:model_training}, we introduced $\theta^\ell$ as a trainable parameter that increases the saliency score of \textsc{[CLS]}. We can deduce from Equations \ref{eq:loss_CP} and \ref{eq:S_hat} that this parameter does not exist in the model's computational DAG and we need to compute the derivative of $\tilde{S}^\ell$ w.r.t. $\theta^\ell$ to train this parameter. Hence, first we assume that $\tilde{S}^\ell$ is a close estimate of $\hat{S}^\ell$ (due to the CPs' training objective). Second, using a dummy variable $\theta^\ell_d$---that is involved in the computational graph and is always equal to $1$---we reformulate $\tilde{S}^\ell$:
\begin{equation}
\label{eq:S_tilde_reformulated} 
\hat{S}^\ell_i\approx\tilde{S}^\ell_i=\frac{\theta^\ell_d \tilde{S}^\ell_1\mathbf{1}[i=1] + \tilde{S}^\ell_i\mathbf{1}[i>1]}{\theta^\ell_d \tilde{S}^\ell_1+\sum_{i=2}^{n}\tilde{S}_i^\ell}
\end{equation}
This reformulation is valid due to $\theta^\ell_d = 1$ and $\sum_{i=1}^{n}\tilde{S}_i^\ell = 1$. Now we compute the partial derivative w.r.t. $\theta^\ell_d$ which is the gradient that is computed in the backpropagation:
\begin{equation}
\frac{\partial\tilde{S}^\ell_i}{\partial\theta^\ell_d} =\frac{\tilde{S}^\ell_1(\sum_{i=2}^{n}\tilde{S}_i^\ell\mathbf{1}[i=1] - \tilde{S}^\ell_i\mathbf{1}[i>1])}{(\theta^\ell_d \tilde{S}^\ell_1+\sum_{i=2}^{n}\tilde{S}_i^\ell)^2}
\end{equation}
By knowing that $\theta^\ell_d = 1$:
\begin{equation}
\frac{\partial\tilde{S}^\ell_i}{\partial\theta^\ell_d} =\tilde{S}^\ell_1((1-\tilde{S}^\ell_1)\mathbf{1}[i=1] - \tilde{S}^\ell_i\mathbf{1}[i>1])
\end{equation}
Now using our initial assumption ($\hat{S}_i^\ell \approx \tilde{S}_i^\ell$), we can substitute $\tilde{S}_i^\ell$ with $\hat{S}_i^\ell$ based on Equation \ref{eq:S_hat}:
\begin{equation}
\label{eq:app_Si_tilde_theta_d}
\begin{aligned}
\frac{\partial\tilde{S}^\ell_i}{\partial\theta^\ell_d} &=\hat{S}^\ell_1((1-\hat{S}^\ell_1)\mathbf{1}[i=1] - \hat{S}^\ell_i\mathbf{1}[i>1]) \\
&= \frac{\theta^\ell S^\ell_1(\sum_{i=2}^{n}S_i^\ell\mathbf{1}[i=1] - S_i^\ell\mathbf{1}[i>1])}{(\theta^\ell S^\ell_1+\sum_{i=2}^{n}S_i^\ell)^2}
\end{aligned}
\end{equation}
In addition, the gradient of $\hat{S}_i^\ell$ w.r.t. $\theta^\ell$ is as follows (cf. Equation \ref{eq:S_hat}):
\begin{equation}
\label{eq:app_Si_tilde_theta}
\frac{\partial\hat{S}^\ell_i}{\partial\theta^\ell} = \frac{ S^\ell_1(\sum_{i=2}^{n}S_i^\ell\mathbf{1}[i=1] - S_i^\ell\mathbf{1}[i>1])}{(\theta^\ell S^\ell_1+\sum_{i=2}^{n}S_i^\ell)^2}
\end{equation}
By comparing Equations \ref{eq:app_Si_tilde_theta_d} and \ref{eq:app_Si_tilde_theta}, these derivatives are related with a term of $\theta^\ell$:
\begin{equation}
\frac{\partial\hat{S}^\ell_i}{\partial\theta^\ell} \approx \frac{\partial\tilde{S}^\ell_i}{\partial\theta^\ell} = \frac{1}{\theta^\ell}\frac{\partial\tilde{S}^\ell_i}{\partial\theta^\ell_d}
\end{equation}
Therefore, during training, we can compute the gradient w.r.t. the dummy variable $\theta^\ell_d$ and then divide it by $\theta^\ell$.
\section{Evaluating PoWER-BERT in Single Instance Mode}
\label{sec:power-bert-single-instance}
Due to the static structure of PoWER-BERT, the speedup ratios reported in \citet{goyal2020power} are based on wall time acceleration with batch-wise inference procedure. This means that some inputs might need extra padding to make all inputs with the same token length. However, since our approach and other dynamic approaches are based on single instance inference, in our procedure inputs are fed without being padded. To even out this discrepancy, we apply a single instance flops computation on the PoWER-BERT, which means we compute the computational cost for all input lengths that appear in the test dataset. Some instnaces may have shorter input length than some values in the resulting retention configuration (number of tokens that are retained in each layer). To overcome this issue, we update the retention configuration by selecting the minimum between the input length and each layers' number of tokens retained, to build a new retention configuration for each input length. For instance, if the retention configuration trained model on a given task be (153, 125, 111, 105, 85, 80, 72, 48, 35, 27, 22, 5), for an input with 75 tokens length, the new configuration which is used for speedup computation will be: (75, 75, 75, 75, 75, 75, 72, 48, 35, 27, 22, 5). 

\section{AdapLeR Training Hyperparameters}
\label{sec:Hyperparams_app}
For the initial step of fine-tuning BERT, we used the hyperparameters in Table \ref{tab:HyperparametersFT}. For both fine-tuning and training with length reduction, we employed an AdamW optimizer \citep{loshchilov2018decoupled} with a weight decay rate of 0.1, warmup proportion 6\% of total training steps and a linear learning rate decay which reaches to zero at the end of training. 

\begin{table}[ht]
\centering
\resizebox{0.99\linewidth}{!}{
\begin{tabular}{l  c c c c} 
\toprule
Dataset & Epoch & LR & MaxLen. & BSZ\\
\midrule

SST-2 & $5$ & $2e$-$5$ & $64$ & $32$\\
IMDB & $5$ & $2e$-$5$ & $512$ & $16$\\
HateXplain & $5$ & $3e$-$5$ & $72$ & $32$\\
MRPC & $5$ & $2e$-$5$ & $128$ & $32$\\
MNLI & $3$ & $2e$-$5$ & $128$ & $32$\\ 
QNLI & $5$ & $2e$-$5$ & $128$ & $32$\\ 
AG's News & $5$ & $2e$-$5$ & $128$ & $32$\\
DBpedia & $3$ & $2e$-$5$ & $128$ & $32$\\
\bottomrule
\end{tabular}
}
\caption{Hyperparameters in each dataset; LR: Learning rate;  BSZ: Batch size;  MaxLen: Maximum Token Length}
\label{tab:HyperparametersFT}
\end{table}

For the adaptive length reduction training step, we also used the same hyperparameters in Table \ref{tab:HyperparametersFT} with two differences: Since MRPC and CoLA have small training sets, to prolong the gradual soft-removal process, we increased the training duration to 10 epochs. Moreover, we increase the learning rate to 3e-5. Other hyperparameters are stated in Table \ref{tab:HyperparametersALR}. To set a trend for $\lambda$, it needs to start from a small but effective value ($10< \lambda < 100$) and grow exponentially per each epoch to reach an extremely high amount at the end of the training to mimic a hard removal function ($1e$+$5 < \lambda$). Hence, datasets with the same amount of training epochs have similar $\lambda$ trends.

\begin{table}[ht]
\centering
\resizebox{0.85\linewidth}{!}{
\begin{tabular}{l  c c r} 
\toprule
Dataset & $\gamma$ & $\phi$ & $\lambda$\\
\midrule

SST-2 & 5e-3 & 5e-4 & $10^{Epoch}$\\
IMDB & 5e-3 & 5e-4 & $10^{Epoch}$\\
HateXplain & 5e-2 & 2e-2 & $50^{Epoch}$\\
MRPC & 3e-2 & 5e-2 & $10\times3^{Epoch}$\\
MNLI & 5e-3 & 5e-4 & $50^{Epoch}$\\ 
QNLI & 5e-3 & 1e-4 & $10^{Epoch}$\\ 
AG's News & 1e-1 & 1e-1 & $10^{Epoch}$\\
DBPedia & 1e-1 & 1e-1 & $50^{Epoch}$\\ 
\bottomrule
\end{tabular}
}
\caption{AdapLeR hyperparameters in each dataset; Since $\lambda$ increases exponentially on each epoch the coorresponding formula is written.}
\label{tab:HyperparametersALR}
\end{table}

\section{Statistics of Datasets}

\begin{table}[ht]
\centering
\resizebox{\columnwidth}{!}{
\begin{tabular}{lccc}
     \toprule
     & \multicolumn{2}{c}{Number of Examples} & Number of Tokens \\ \cmidrule(lr){2-3} Task & Train & Test & Mean / Median\\
     \midrule
     SST-2 & 67349 & 1821 & 14 / 11\\
     IMDB & 25000 & 25000 & 275 / 233\\
     HateXplain & 15383 & 1924 & 30 / 27\\
     MRPC & 3668 & 1725 & 53 / 53\\
     MNLI & 392702 & 9796$^\dagger$ / 9847$^\ddagger$ & 40 / 37\\
     QNLI & 104743 & 5463 & 50 / 47\\
     AG's News & 120000 & 7600 &  53 / 51\\
     DBPedia & 560000 & 70000 & 64 / 64\\
     \bottomrule
\end{tabular}
}
\caption{The statistics of datasets: number of training and test examples and average and median of sequence length (number of tokens) of test examples based on BERT's tokenizer. $^\dagger$ and $^\ddagger$ indicate \emph{matched} and \emph{mismatched} versions of MNLI test split, respectively.}
\label{tab:data_statitics}
\end{table}

\section{Additional Qualitative Examples}
\begin{figure*}[!t]
\centering
    \includegraphics[width=0.85\linewidth]{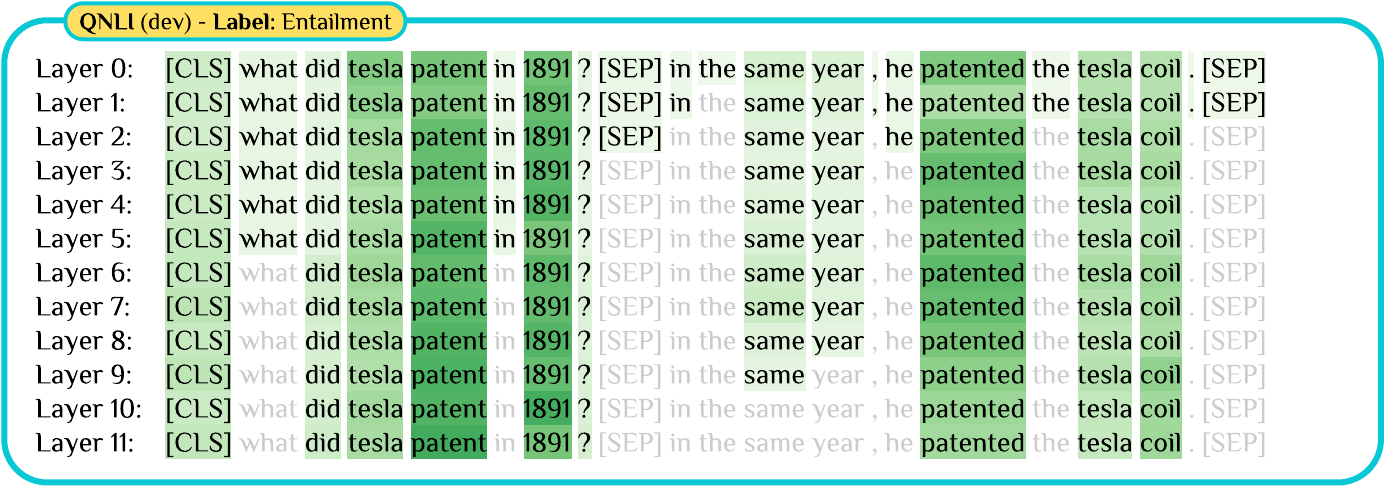}
    \includegraphics[width=0.9\linewidth]{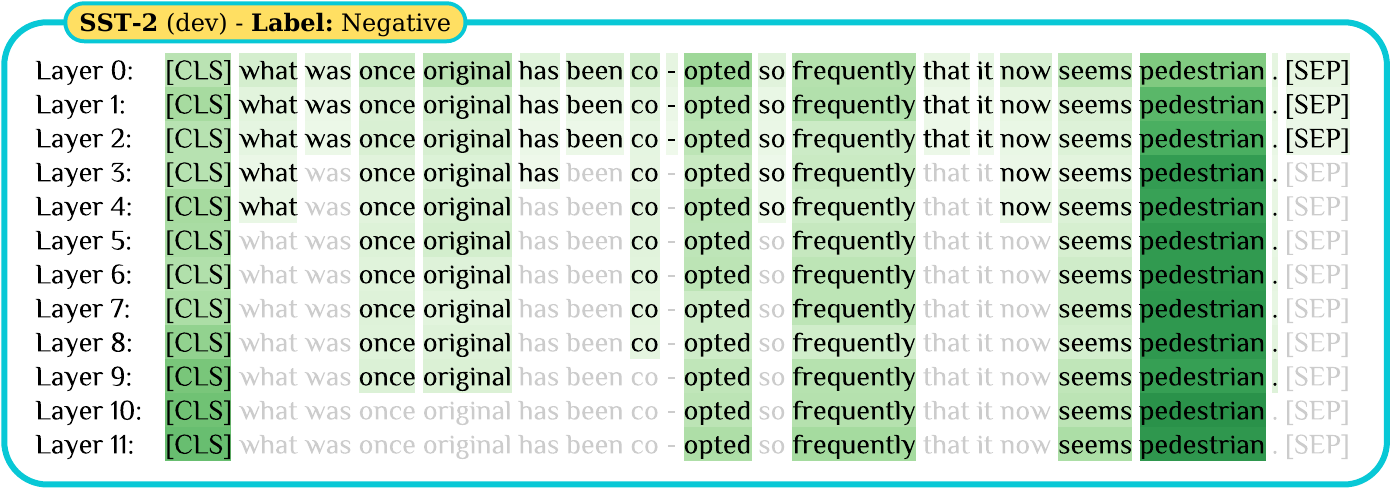}
    \includegraphics[width=\linewidth]{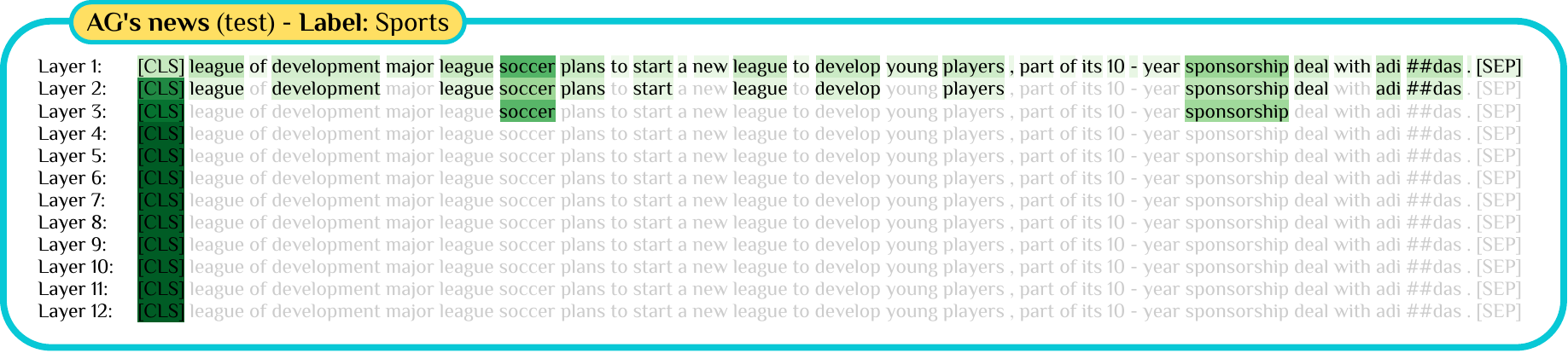}
    \caption{The illustration of contribution scores obtained by CPs in each layers of the model for different input examples from QNLI (Question-answering NLI), SST-2 (sentiment), and AG's news (topic classification) tasks. The color intensity indicates the degree of contribution scores. Only the highlighted token representations are processed in each layer}
    \label{fig:additional_qualitative_analysis}
\end{figure*}

\end{document}